\documentclass[letterpaper, 10 pt, conference]{ieeeconf}  

\IEEEoverridecommandlockouts                              

\usepackage{graphics} 
\usepackage{epsfig} 
\usepackage{cite}
\usepackage{times} 
\usepackage{amsmath} 
\usepackage{amssymb}  
\usepackage{graphicx}
\usepackage{mathrsfs}   
\usepackage{bm} 
\usepackage{caption}
\usepackage{subcaption}
\usepackage{url}
\usepackage{color}
\usepackage{algorithm}
\usepackage{algorithmic}

\usepackage{amsthm}

\usepackage{etoolbox}
\AtBeginEnvironment{align}{\setcounter{subeqn}{0}}
\newcounter{subeqn} \renewcommand{\thesubeqn}{\theequation\alph{subeqn}}%
\newcommand{\subeqn}{%
  \refstepcounter{subeqn}
  \tag{\thesubeqn}
}

\newtheorem{Assumption}{Assumption}

\newtheorem{theorem}{Theorem}
\newtheorem*{opt}{Repair Optimization Problem (R-OPT)}
\newtheorem{remark}[theorem]{Remark}
\newtheorem{definition}{Definition}

\newcommand{\problem}{Minimal Repair Problem}

\newcommand{\ts}{\textsuperscript}

\title{\LARGE \bf
Local Repair of Neural Networks Using Optimization
}
\author{Keyvan Majd$^{1}$, Siyu Zhou$^{1}$, Heni Ben Amor$^{1}$, Georgios Fainekos$^{1}$, and Sriram Sankaranarayanan$^{2}$ 
\thanks{This research was partially funded by NSF CNS 1932068 and 1932189, and DARPA AMP N6600120C4020.}
\thanks{$^{1}$ K.Majd, S. Zhou, H. Ben Amor, and G. Fainekos
        {\tt\small \{majd,szhou22,hbenamor,fainekos\}@asu.edu} are with SCAI, Arizona State University, Tempe, AZ, USA.}%
\thanks{$^{2}$ S. Sankaranarayanan
        {\tt\small \{first.lastname\}@colorado.edu} is with the Department of Computer Science at University of Colorado Boulder, Boulder, CO, USA.}
}

\begin{document}

\maketitle
\thispagestyle{empty}
\pagestyle{empty}

\begin{abstract}
In this paper, we propose a framework to repair a pre-trained feed-forward neural network (NN) to satisfy a set of properties. We formulate the properties as a set of predicates that impose constraints on the output of NN over the target input domain. We define the NN repair problem as a Mixed Integer Quadratic Program (MIQP) to adjust the weights of a single layer subject to the given predicates while minimizing the original loss function over the original training domain. We demonstrate the application of our framework in bounding an affine transformation, correcting an erroneous NN in classification, and bounding the inputs of a NN controller.  
\end{abstract}

\section{Introduction}

Artificial Neural Networks (NN) have shown potential in enabling autonomy for tasks that are hard to formally define or specify.
For instance, NN can be used to learn and replicate multi-robot behaviors \cite{RiviereEtAl2020ral,ZhouEtAl2019iros,FanLLP2020ijrr}, or to model complex aircraft collision avoidance protocols \cite{JulianEtAl2016dasc}.
Driving is another example of such a task where not all driving situations can be predicted in advance and, therefore, it may be beneficial to extrapolate behaviors from previously learned scenarios.
Thus, learning for driving \cite{PanEtAl2018rss,KuuttiEtAl2021its,StricklandFBA2018icra} is gaining ground as a possible solution to the automated driving or driver assistance challenge problems.

Nevertheless, NN-based autonomous systems may produce unsafe behaviors when they encounter inputs in regions where an insufficient number of training samples was available.
Multiple methods exist that attempt to discover such unsafe behaviors (also known as adversarial samples or counter-examples) for autonomous systems  \cite{TianEtAl2018icse,Dreossi2019,TuncaliEtAl2020tiv}.
Other methods investigate classes of systems and NN for which provable properties can be established \cite{SunKS2019hscc,IvanovEtAl2020,DuttaEtAl2018adhs,TjengXT2019iclr}.

In either case, an open challenge is how to repair or modify NN-based systems when they are found  to not be safe.  
In the case of testing based methods (and some verification methods), adversarial samples are returned that can be added to the training set to improve the accuracy of the NN \cite{YaghoubiF2019tecs,DreossiJS2018arxiv,DreossiGYKSS2018ijcai}.
However, adding multiple adversarial samples to the training set might produce over-fitting, or even ``oscillations", i.e., adding training samples from one region may result in reduced accuracy in another region.
On the other hand, many verification methods may not even return any information when an NN-based system fails to satisfy its safety requirements.
Hence, repairs may not be as straightforward as when counterexamples are available.

In this paper, we make progress toward the repair problem for NN-based systems.
We pose the following question: given a set of safety constraints on the output of the NN component, is it possible to modify the NN such that the safety constraints are satisfied while the NN accuracy over the original training set remains high?
We propose an optimization based solution by posing the repair problem as a Mixed-Integer Quadratic Program (MIQP).
Our framework is presented on a number of challenging problems relevant to robotics: forward kinematics of arm manipulators, aircraft collision avoidance, and mobile robot motion control.

The repair problem \cite{goldberger2020minimal,DongEtAl2021arxiv,CruzFS2021arxiv} has objectives that typically extend beyond just identifying adversarial samples to be added to the training set \cite{YaghoubiF2019tecs,DreossiJS2018arxiv,DreossiGYKSS2018ijcai}.
In \cite{CruzFS2021arxiv}, the authors address the repair problem for NN-based controllers with a Two-Level Lattice (TLL) architecture.
The main step in the framework is a local repair to the controller which is only activated in the states in the vicinity of any adversarial samples. 
A condition for applying the repair is that any discovered unsafe regions do not overlap with previously trained regions.
A limitation of the method in \cite{CruzFS2021arxiv} is that it is only applicable to the NN architecture. 
In contrast, our approach is applicable to general feed forward NN and there are no conditions on the adversarial samples. 

The repair process in \cite{DongEtAl2021arxiv} is executed in the loop with a verification engine. 
Given an adversarial sample, the neighborhood around it is partitioned until a region is identified where for every point in that region the safety constraints are violated.
Then, the neurons that cause the violation are identified and they are repaired.
Even though the method applies to general feedforward NN, the repair formulation does not capture the effect of the neuron modifications to the performance of the NN overall.

Finally, the authors in \cite{goldberger2020minimal} propose a verification-based method for minimally modifying the network weights so as to satisfy a given property. 
Due to the use of verification, they guarantee the correctness of the repaired model. 
However, this work restricts the repair to the last layer and only guarantees the minimality of the weight changes. 
Their method is also limited to guaranteed modification of only one adversarial input.



The following are the main contributions of this paper. 
First, we propose a layer-wise neural network repair framework using Mixed-integer Quadratic Programming (MIQP) to satisfy constraints on the output of the network for a given input domain (Sec. \ref{sec: MinRepair-MIP}).  
Second, while our method repairs the network for the specified local input domain, it maintains the prediction/classification accuracy of the network over its original trained space. 
Finally, we demonstrate the application of our method in addressing safety, model improvement, and verification problems in a variety of robot learning domains (Sec. \ref{sec: sim-sec}). 

\section{Problem Statement}
\subsection{Notation}

\begin{figure}[t]
    \centering
    \includegraphics[scale=0.27]{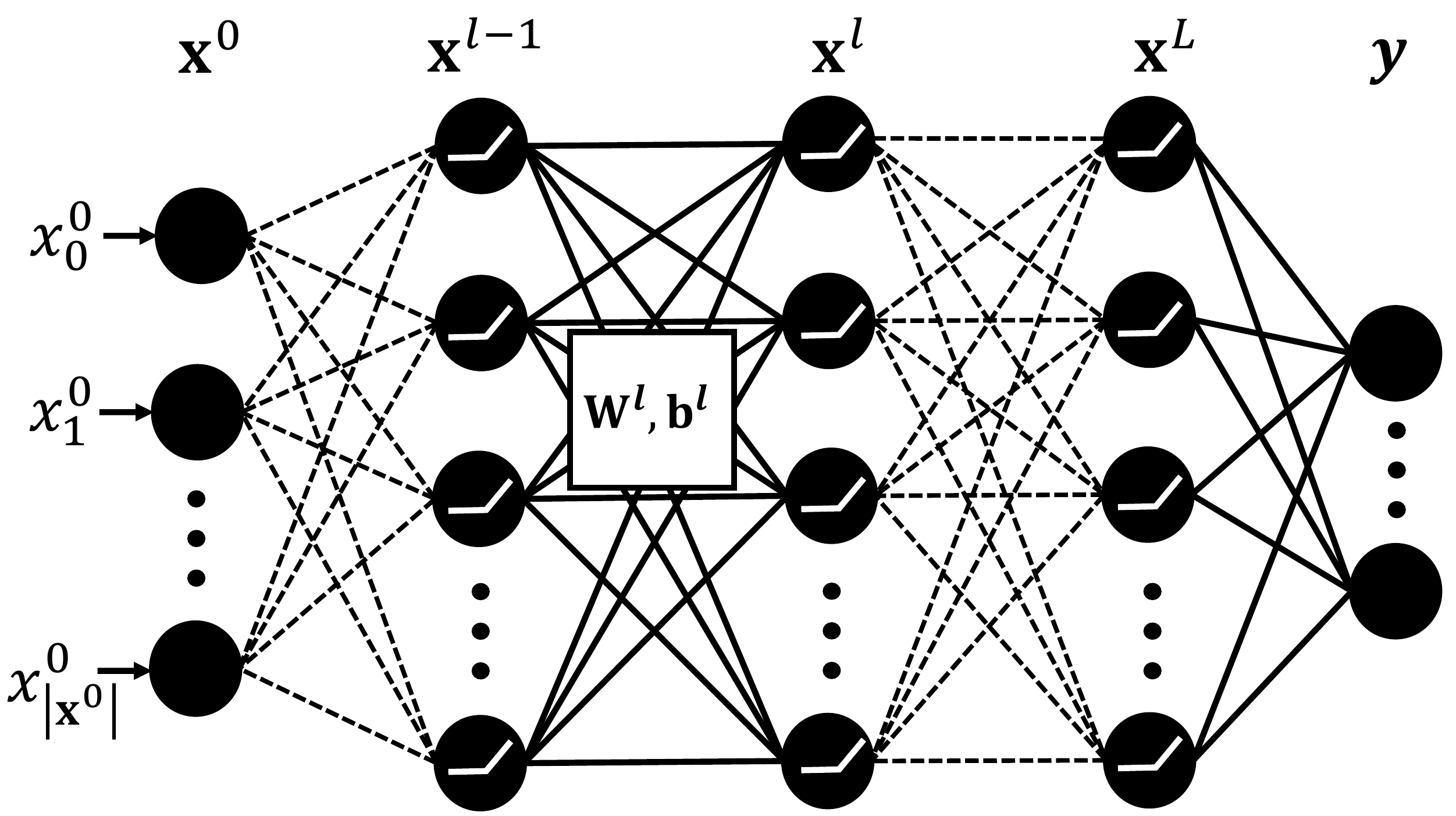}
    \caption{Multi-layer Perceptron}
    \label{fig:MLP}
\end{figure}

We denote the set of variables $a_1,a_2,\cdots,a_N$ with $\{a_n\}^N_{n=1}$.
Let $N\!N$ be a neural network with $L$ hidden layers, as shown in Fig. \ref{fig:MLP}. 
The nodes values at each layer $l \in \{l_i\}^L_{l=0}$ are represented by a vector $\mathbf{x}^l=[x^l_0,x^l_1,\cdots,x^l_{\lvert\mathbf{x}^l\rvert}]^T$, where $\lvert\mathbf{x}^l\rvert$ denotes the dimension of layer $l$. The network output is also denoted as $\mathbf{y}$.
Each two subsequent layers are fully connected with weighted edges and bias terms. 
The notation 
\[ \mathbf{x}^{l-1}\xrightarrow{\mathbf{W}^l,\mathbf{b}^l} \mathbf{x}^l \]  
represents the weight matrix $\mathbf{W}^l$ and the bias vector $\mathbf{b}^l$ of layer $l$ connecting the vectors $\mathbf{x}^{l-1}$ to $\mathbf{x}^l$ with 
$$
\mathbf{W}^l = \begin{bmatrix}
w^l_{00} & w^l_{10} &\cdots & w^l_{\lvert\mathbf{x}^{l-1}\rvert0}\\
w^l_{01} & w^l_{11} &\cdots & w^l_{\lvert\mathbf{x}^{l-1}\rvert1}\\
\vdots   & \vdots   &   \ddots    &\vdots\\
w^l_{0\lvert\mathbf{x}^l\rvert} & w^l_{1\lvert\mathbf{x}^l\rvert} &\cdots & w^l_{\lvert\mathbf{x}^{l-1}\rvert\lvert\mathbf{x}^l\rvert}
\end{bmatrix}, 
\mathbf{b}^l = \begin{bmatrix}
b^l_0\\
b^l_1\\
\vdots\\
b^l_{\lvert\mathbf{x}^l\rvert}
\end{bmatrix}.
$$
The training data set of $N$ inputs $\mathbf{x}^0_n$ and target outputs $\mathbf{t}_n$ is denoted by $\{(\mathbf{x}^0_n,\mathbf{t}_n)\}^N_{n=1} \subseteq \mathcal{X}_o\times\mathcal{T}\subseteq\mathbb{R}^{\lvert\mathbf{x}^0\rvert}\times\mathbb{R}^{\lvert\mathbf{t}\rvert}$.
We use $x^l_{j,n}$ to denote the $j$th node value of layer $l$ for sample $n$ and $\mathbf{x}^l_n$ to denote the vector of nodes at layer $l$ for sample $n$. 
The value of each hidden node is calculated using the weighted sum of the nodes in its previous layer passed through a nonlinear activation function. 

In this work, we solve the repair problem for networks that use the Rectified Linear Unit (ReLU) activation function $R(z) = \max\{0,z\}$. 
Thus, given the $n$th sample, in the hidden layer $l:\mathbf{x}^{l-1}\xrightarrow{\mathbf{W}^l,\mathbf{b}^l}
\mathbf{x}^l$, we have $\mathbf{x}^l = R\left(\mathbf{W}^l\mathbf{x}^{l-1}+\mathbf{b}^l\right)$. 
An activation function is not applied to the last layer, so $\mathbf{y} = \mathbf{W}^{L+1}\mathbf{x}^{L}+\mathbf{b}^{L+1}$.


\subsection{Problem Formulation}
\begin{definition}[Repair Problem] Let $N\!N_o$ be a trained neural network over the training input-output space $\mathcal{X}_o\times\mathcal{T}\subseteq\mathbb{R}^{\lvert\mathbf{x}^0\rvert}\times\mathbb{R}^{\lvert\mathbf{t}\rvert}$ and $\Psi(\mathbf{y},\mathbf{x}^0)$ be a constraint (predicate) on the output $\mathbf{y}$ of $N\!N_o$ for a set of inputs of interest $\mathbf{x}^0\in \mathcal{X}_r\subseteq\mathcal{X}_o$. The Repair Problem is to modify the weight and bias terms of $N\!N_o$ such that $\Psi(\mathbf{y},\mathbf{x}^0)$ holds for the new neural network $N\!N_r$.
\end{definition}

By repairing $N\!N_o$, we also aim to maintain the performance of the original network. 
Thus, the repaired network $N\!N_r$ should both satisfy $\Psi(\mathbf{y},\mathbf{x}^0)$ for the inputs $\mathbf{x}^0\in\mathcal{X}^{r}$ and maintain the prediction/classification performance of $N\!N_o$ for the inputs in the original training space domain $\mathcal{X}_o\times\mathcal{T}$. 
To satisfy the latter, the minimal network modification method proposed in \cite{goldberger2020minimal} minimizes the $d$-norm error between each two weight and bias vectors of $N\!N_o$ and $N\!N_r$ while satisfying $\Psi(\mathbf{y},\mathbf{x}^0)$. 
Although the authors in \cite{goldberger2020minimal} showed that their method can successfully modify the network to correct the observed misclassified points in a classification problem, a minimal deviation from the original weights is not necessarily a sufficient guarantee for maintaining the prediction/classification performance of the original network over the whole input space $\mathcal{X}_r$. 
As an example, assume $NN_o$ is trained as a binary classifier (the highest output determines the classifier label for a given input) with one hidden layer ($L=1$) and each layer of size two. 
Assume the original weights and biases are 
\begin{align*}
\mathbf{W}^1&= \begin{bmatrix}
0.2 & 1 \\
0.01 & 1
\end{bmatrix}~, 
\mathbf{b}^1 = \begin{bmatrix}
0.2\\
-0.11
\end{bmatrix}, \\
\mathbf{W}^2&= \begin{bmatrix}
0.5 & 1 \\
2 & -1
\end{bmatrix}, 
\mathbf{b}^2 = \begin{bmatrix}
-0.2\\
-1
\end{bmatrix}.
\end{align*}
Assume $\mathbf{x}^0=[9,0]^T$ is found as a misclassified point in $N\!N_o$ for which the true label is $y_0>y_1$. 
A minimal change of weight $w^1_{01} = 0.01$ to $0.02$ can repair this network to correctly classify this point. 
However, this repair changes the label of the classifier for $\mathbf{x}^0=[100,20]^T$ and creates a new misclassified point that was correctly classified in the original network. 
Hence, a subtle change in the weights may cause the network to significantly deviate from its original performance. Therefore, it is important for the repaired network $N\!N_r$ to also minimize the defined loss function $E(\mathbf{W},\mathbf{b})$ of the original network $N\!N_o$.

\begin{definition}[Minimal Repair Problem] \label{MinRepair-formulation} Let $N\!N_o$ be a trained neural network over the training input-output space $\mathcal{X}_o\times\mathcal{T}\subseteq\mathbb{R}^{\lvert\mathbf{x}^0\rvert}\times\mathbb{R}^{\lvert\mathbf{t}\rvert}$ and $\Psi(\mathbf{y},\mathbf{x}^0)$ be a predicate that represents some constraints on the output $\mathbf{y}$ of $N\!N_o$ for the set of inputs of interest $\mathbf{x}^0\in \mathcal{X}_r\subseteq\mathcal{X}_o$. 
The Minimal Repair Problem is to modify the weights of $N\!N_o$ such that the new network $N\!N_r$ satisfies $\Psi(\mathbf{y},\mathbf{x}^0)$ while preserving the topology of $N\!N_o$ and minimizing the original defined loss function $E(\mathbf{W},\mathbf{b})$ over the original domain $\mathcal{X}_o\times\mathcal{T}$.
\end{definition}

\section{Layer-wise Network Repair as a Mixed-Integer Quadratic Program (MIQP)} \label{sec: MinRepair-MIP}
To solve the \problem, we can formulate it as an optimization to minimize the loss function $E(\mathbf{W},\mathbf{b})$ subject to $(\mathbf{x}^0,\mathbf{t})\in\mathcal{X}_o\times\mathcal{T}$ and $\Psi(\mathbf{y},\mathbf{x}^0)$ for $\mathbf{x}^0\in \mathcal{X}^{r}$. 
However, the resulting optimization problem is non-convex and difficult to solve due to the nonlinear ReLU activation function and high-order nonlinear constraints resulted from the multiplication of terms involving the weight/bias variables. 
An alternative approach is to obtain a sub-optimal solution by just focusing on a single layer repair. 
Layer-wise repair just modifies the weights and biases of a single layer to adjust the predictions so as to minimize $E(\mathbf{W},\mathbf{b})$ and to satisfy $\Psi(\mathbf{y},\mathbf{x}^0)$. 
The layer-wise Minimal Repair Problem is defined as follows
\begin{definition}[Layer-wise Minimal Repair Problem] \label{SingleLayMinRepair-formulation} Let $N\!N_o$ denote a trained neural network with $L$ hidden layers over the training input-output space $\mathcal{X}_o\times\mathcal{T}\subseteq\mathbb{R}^{\lvert\mathbf{x}^0\rvert}\times\mathbb{R}^{\lvert\mathbf{t}\rvert}$ and $\Psi(\mathbf{y},\mathbf{x}^0)$ denote a predicate that represents some constraints on the output $\mathbf{y}$ of $N\!N_o$ for the set of inputs of interest $\mathbf{x}^0\in \mathcal{X}_r\subseteq\mathcal{X}_o$. 
The Layer-wise Minimal Repair Problem is to modify the weights of a layer $l\in\{1,\cdots,L+1\}$ in $N\!N_o$ such that the new network $N\!N_r$ satisfies $\Psi(\mathbf{y},\mathbf{x}^0)$ while preserving the topology of $N\!N_o$ and minimizing the original defined loss function $E(\mathbf{W}^l,\mathbf{b}^l)$ over the original domain $\mathcal{X}_o\times\mathcal{T}$.
\end{definition}

In our optimization problem, we modeled the ReLU activation function $R(z)=\max\{0,z\}$ with  a disjunctive inequality constraint \cite{balas1979disjunctive}. 
Assume $\phi$ is a Boolean variable and $z = x-s$, where $x,s\geq 0$. 
In the disjunctive inequality (\ref{eq: disjunction}), we have $x=z$ when $\phi=1$ ($z\geq 0$) and $x=0$ when $\phi=0$ ($z=-s< 0$). Here $x$ determines the output of ReLU activation function $R(z)$,
\begin{align}\label{eq: disjunction}
    \left[\begin{array}{c}
         \phi = 0\\
          x\leq 0
    \end{array}\right]\bigvee\left[\begin{array}{c}
         \phi = 1\\
          s\leq 0
    \end{array}\right]. 
\end{align}
Disjunctive Ineq. (\ref{eq: disjunction}) can be relaxed into mixed integer algebraic equations using the Big-M or Convex-Hull reformulation \cite{tsay2021partition,belotti2011disjunctive}. 
We also assume that the predicate $\Psi(\mathbf{x}^0,\mathbf{y})$ is represented as a linear constraint.
\begin{Assumption}\label{assump: linear_predicate} 
The predicates $\Psi(\mathbf{x}^0,\mathbf{y})$ are formulated as linear equality/inequality constraints.
\end{Assumption}

\begin{figure}[t]
    \centering
    \includegraphics[scale=0.29]{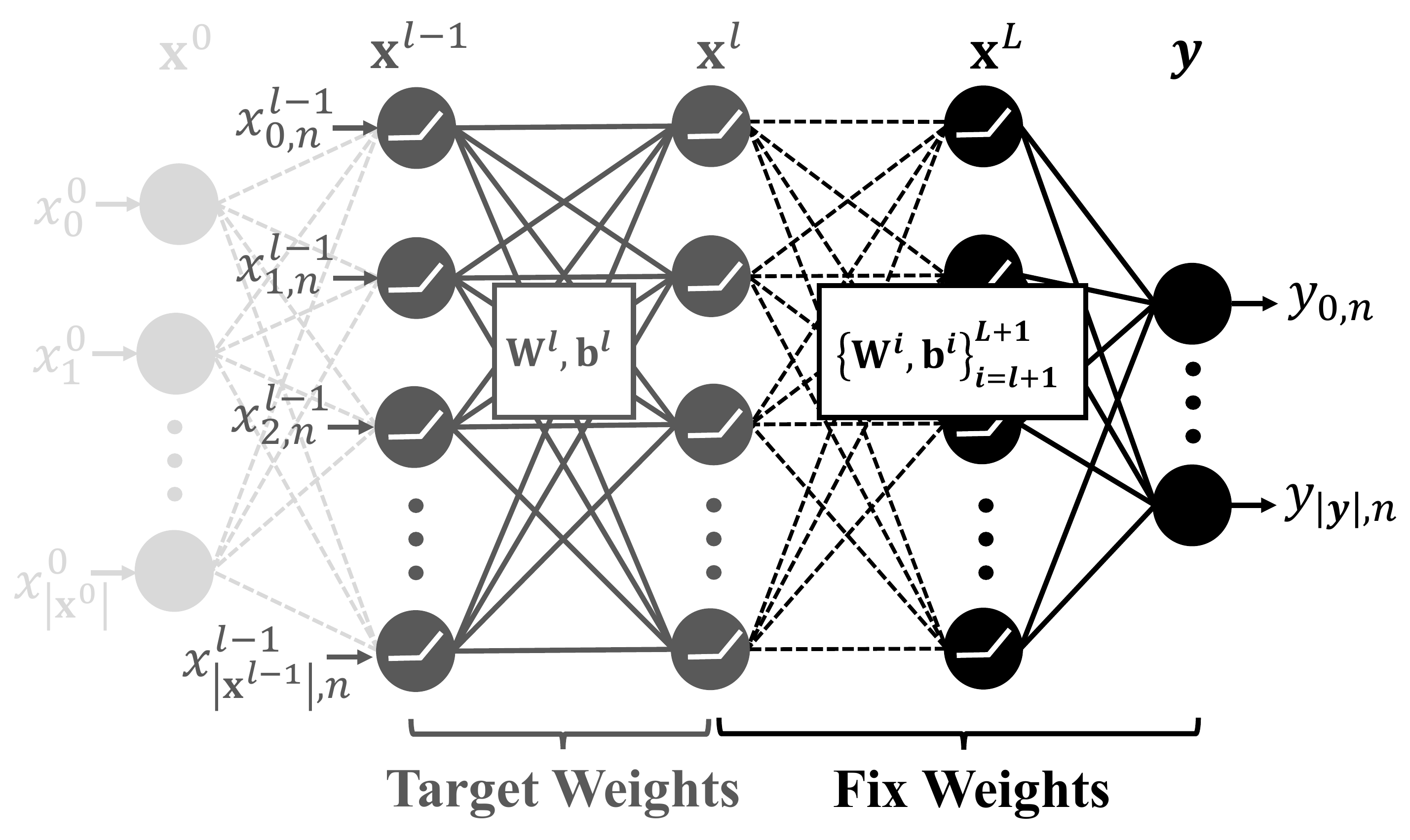}
    \caption{Multi-layer Perceptron repair given a sample $n$}
    \label{fig:MLP-rpair}
\end{figure}

Moreover, since $\mathcal{X}_r$ and $\mathcal{X}_o$ are not necessarily convex, we formulate the layer-wise minimal optimization problem over a data set sampled from $\mathcal{X}_o\times\mathcal{T}$ and $\mathcal{X}_r\times \Tilde{{\mathcal{T}}}$, where $\Tilde{{\mathcal{T}}}$ is the set of original target values of inputs in $\mathcal{X}_r$. 
\begin{remark}
The predicate $\Psi(\mathbf{x}^0,\mathbf{y})$ defined over $\mathbf{x}^0\in \mathcal{X}_r$ is not necessarily compatible with the target values in $\Tilde{\mathcal{T}}$. 
It means that the predicate may bound the NN output for $\mathcal{X}_r$ input space such that not allowing an input $\mathbf{x}^0\in \mathcal{X}_r$ to reach its target value in $\Tilde{\mathcal{T}}$. 
It is a natural constraint in many applications. 
For instance, due to the safety constraints, we may not allow a NN controller to follow its original control reference for a given unsafe set of input states. 
\end{remark}
For a given layer $l$, we also define $E(\mathbf{W}^l,\mathbf{b}^l)$ in the form of sum of square loss as follows 
\begin{align}\label{eq: loss}
    E(\mathbf{W}^l,\mathbf{b}^l) = \sum^{N}_{n=1}\lVert \mathbf{y}_n(\mathbf{x}_n^0,\mathbf{W}^l,\mathbf{b}^l)-\mathbf{t}_n\rVert^2_2,
\end{align}
where $\mathbf{W}^l$ and $\mathbf{b}^l$ are the weight and bias vectors of the layer $l$, respectively, and $\{(\mathbf{x}^0_n,\mathbf{t}_n)\}^N_{n=1}\sim \mathcal{X}_o\times\mathcal{T} \cup \mathcal{X}_r\times\Tilde{\mathcal{T}}$ ($\lVert\cdot\rVert_2$ denotes the Euclidean norm).
Here, since we only repair the weights and biases of the target layer $l$, the loss term $E$ is a function of the weight matrix and bias vector of $\mathbf{W}^l$ and $\mathbf{b}^l$, respectively. 
Hence, the weight and bias terms of all layers except the target layer $l$ are fixed, as shown in Fig. \ref{fig:MLP-rpair}. 
Considering ReLU activation functions formulated by disjunctive Ineq. (\ref{eq: disjunction}), the loss function (\ref{eq: loss}) and Asmp. \ref{assump: linear_predicate}, we formulate the Layer-wise Minimal Repair Problem in Def. \ref{SingleLayMinRepair-formulation} as a Mixed Integer Quadratic Program (MIQP).

\begin{opt} \label{opt: minrepair1}
Let $N\!N_o$ be a neural network with $L$ hidden layers, $\Psi(\mathbf{y},\mathbf{x}^0)$ be a predicate, and $\{(\mathbf{x}^0_n,\mathbf{t}_n)\}^N_{n=1}$ be an input-output data set sampled from  $(\mathcal{X}_o\times\mathcal{T}) \cup (\mathcal{X}_r\times\Tilde{\mathcal{T}})$ over the sets $\mathcal{X}_o$, $\mathcal{X}_r$, $\mathcal{T}$, and $\Tilde{\mathcal{T}}$ all as defined in Def. \ref{SingleLayMinRepair-formulation}. Also, let  $E(\mathbf{W}^l,\mathbf{b}^l)$ be the loss function defined in (\ref{eq: loss}). Repair the weight and bias vectors $\mathbf{W}^l$ and $\mathbf{b}^l$ of the $l$\ts{th} layer in $N\!N_o$ (as shown in Fig. \ref{fig:MLP-rpair}) so as to minimize the cost function (\ref{eq:cost}) subject to the constraints (\ref{eq: repair weights-last-layer})-(\ref{eq: weights bound-mid-layer-delta}).

\begin{align}
       & \label{eq:cost}
       \underset{\substack{
       \mathbf{W}^l,\mathbf{b}^l,\mathbf{y}_n,\delta,\\ \{\mathbf{x}_n^i\}_{i=l}^{L},
       \{\mathbf{s}^i\}_{i=l}^{L},\{\bm{\phi}_n^i\}_{i=l}^{L}}
       }
       {\text{min}}~E(\mathbf{W}^l,\mathbf{b}^l)+\delta \\
       & \text{s.t.}\nonumber\\ 
       &~~~\label{eq: repair weights-last-layer}\mathbf{y}_n=\mathbf{W}^{L+1}\mathbf{x}^{L}_n+\mathbf{b}^{L+1},~~~~~~~~~~~~~~ \text{for }\{n\}_{n=1}^N,\\[3pt]
       &~~~\label{eq: repair weights-mid-layer}\mathbf{x}^i_n-\mathbf{s}^i_n=\mathbf{W}^i\mathbf{x}^{i-1}_n+\mathbf{b}^i,\nonumber\\
       &~\qquad\qquad\qquad\qquad\qquad~~~ \text{for } \{n\}_{n=1}^N \text{ and }\{i\}_{i=l}^{L},\refstepcounter{equation}\subeqn\\[3pt]
       &~~~\mathbf{x}^i_n,\mathbf{s}^i_n\geq 0,\qquad\quad~~~~~~~ \text{for } \{n\}_{n=1}^N \text{ and }\{i\}_{i=l}^{L},\subeqn\\[3pt]
       &~~~\bm{\phi}^i_n\in \{0,1\}^{\lvert\mathbf{x}^i\rvert},\quad~~~~~~~ \text{for } \{n\}_{n=1}^N \text{ and }\{i\}_{i=l}^{L},\subeqn\\[3pt]
       &~~~\left[\begin{array}{c}
                    \phi^i_{j,n} = 0\\
                    x^i_{j,n}\leq 0
                    \end{array}\right]\bigvee\left[\begin{array}{c}
                    \phi^i_{j,n} = 1\\
                    s^i_{j,n}\leq 0
                    \end{array}\right],\nonumber\\
       &~\qquad\qquad\qquad~~~~ \text{for } \{n\}_{n=1}^N,~\{i\}_{i=l}^{L}\text{ and }\{j\}_{j=0}^{\lvert\mathbf{x}^i\rvert},\subeqn\\[3pt]
       &~~~\label{eq: requirement-mid-layer}\Psi(\mathbf{y}_n,\mathbf{x}^0_n), \quad~~~~~~~~~~ \text{for }\{n\}_{n=1}^N \text{ and }\mathbf{x}^0_n\in \mathcal{X}_r,\\[3pt]
       &~~~\label{eq: weights bound-mid-layer}\lvert w^l_{kj}-w^{l,prev}_{kj}\rvert\leq\delta,~~~ \text{for } \{k\}_{k=0}^{\lvert\mathbf{x}^{l-1}\rvert}\text{ and }\{j\}_{j=0}^{\lvert\mathbf{x}^l\rvert},\refstepcounter{equation}\subeqn\\[3pt]
       &~~~\lvert b^l_{j}-b^{l,prev}_{j}\rvert\leq\delta,\qquad\quad\quad~~~~~~~~~~~ \text{for } \{j\}_{j=0}^{\lvert\mathbf{x}^l\rvert},\subeqn\\[3pt]
       &~~~\label{eq: weights bound-mid-layer-delta}\delta\geq 0.\subeqn
\end{align}
\end{opt}

The layer-wise repair scheme is shown in Fig. \ref{fig:MLP-rpair} for repairing $\mathbf{x}^{l-1}\xrightarrow{\mathbf{W}^l,\mathbf{b}^l} \mathbf{x}^l$. The sample values of $\mathbf{x}_n^{l-1}$ can be obtained by the weighted sum of the nodes in its previous layers starting from $\mathbf{x}_n^0$ for all $N$ samples $\{n\}^N_{n=1}$. In R-OPT, constraint (\ref{eq: repair weights-last-layer}) represents the forward pass of last layer $\mathbf{x}^{L}\xrightarrow{\mathbf{W}^{L+1},\mathbf{b}^{L+1}} \mathbf{y}$. Constraint 
(5) 
represents the forward pass of layers $\mathbf{x}^{i-1}\xrightarrow{\mathbf{W}^{i},\mathbf{b}^{i}} \mathbf{x}^{i}$ for $\{i\}_{i=l}^{L}$ with ReLU activation function formulated as disjunctive functions (\ref{eq: disjunction}). Constraint (\ref{eq: requirement-mid-layer}) is the given predicate on $\mathbf{y}$ defined over $\mathbf{x}^0\in\mathcal{X}_r$. Finally, Constraint 
(7)
bounds the error between the weight and bias terms $\mathbf{W}$ and $\mathbf{b}$, and their corresponding original weight and bias terms $\mathbf{W}^{prev}$ and $\mathbf{b}^{prev}$ by $\delta$. By adding $\delta$ to (\ref{eq:cost}), we aim to minimize weights deviation as well as $E(\mathbf{W}^l,\mathbf{b}^l)$ in the repair process. 

\begin{remark}
Since the output layer is not passed through a ReLU activation function, applying repair to the last layer $L+1$ does not create  mixed integer constraints 
(5). Therefore, Minimal Repair Problem for the last layer is a Quadratic Program (QP).
\end{remark}

\section{Empirical Experiments}\label{sec: sim-sec}
We show the application of the Layer-wise Minimal Repair framework in bounding an affine transformation, applying constraints on a learned forward kinematics model, correcting an erroneous NN in a classification problem, and applying safety constraints on a NN controller. In all experiments, except the classification problem, we first train a neural network with training data uniformly sampled from the input-output training domain $\mathcal{X}_o\times\mathcal{T}$. We use our proposed repair formulation R-OPT to repair the network given the target layer $l$ and the defined predicate $\Psi(\mathbf{y},\mathbf{x}^0)$ over the input space $\mathcal{X}_r\subseteq\mathcal{X}_o$. In network repair, we use equal samples from both $\mathcal{X}_o\times\mathcal{T}$ and $\mathcal{X}_r\times\Tilde{\mathcal{T}}$. We formulate the optimization using Pyomo open-source optimization language in Python \cite{hart2011pyomo,bynum2021pyomo}, and Pyomo.GDP for modeling the disjunctive constraints \cite{chen2018pyomo}. We use Gurobi 9.1 as solver \cite{gurobi}. 

\subsection{In-place Rotation} 

Our method is first tested on a network that learned in-place rotation. 
We first train a two-hidden-layer network $N\!N_o$ ($L=2$) with $3$ nodes in each layer to learn an in-place 45-degrees-counterclockwise rotation function $\textsc{Rot}\!:\mathcal{X}_o\rightarrow\mathcal{T}$ in 2D space. 
The original input space is $\mathcal{X}_o = \{\mathbf{x}^0=[x,y]\in \mathbb{R}^2\;|\; 1\leq x,y\leq4\}$. The output space $\mathcal{T}$ represents the set of counterclockwise rotated points in $\mathcal{X}_o$ by $45$ degrees defined as $\mathcal{T}=\{\mathbf{y}\in\mathbb{R}^{2}\;|\;\mathbf{y}=\textsc{Rot}(\mathbf{x}_0)\text{ for }\mathbf{x}_0\in\mathcal{X}_o\}$. We repair $N\!N_o$ to push all the outputs $\mathbf{y}$ inside a Quadrilateral defined as
\begin{align*}
    \Psi(\mathbf{y},\mathbf{x}^0)=\{\mathbf{y}\in\mathcal{T}\;|\;\lVert[x,y] - [2.5,2.5]\rVert_1\leq \frac{5\sqrt{2}}{4}\},
\end{align*}
where $\lVert\cdot\rVert_1$ denotes the Taxicab norm. Here, $\mathcal{X}_r=\mathcal{X}_o$. We repair the network with $N=200$ samples.

\begin{figure}[t]
    \centering
    \includegraphics[scale=0.26]{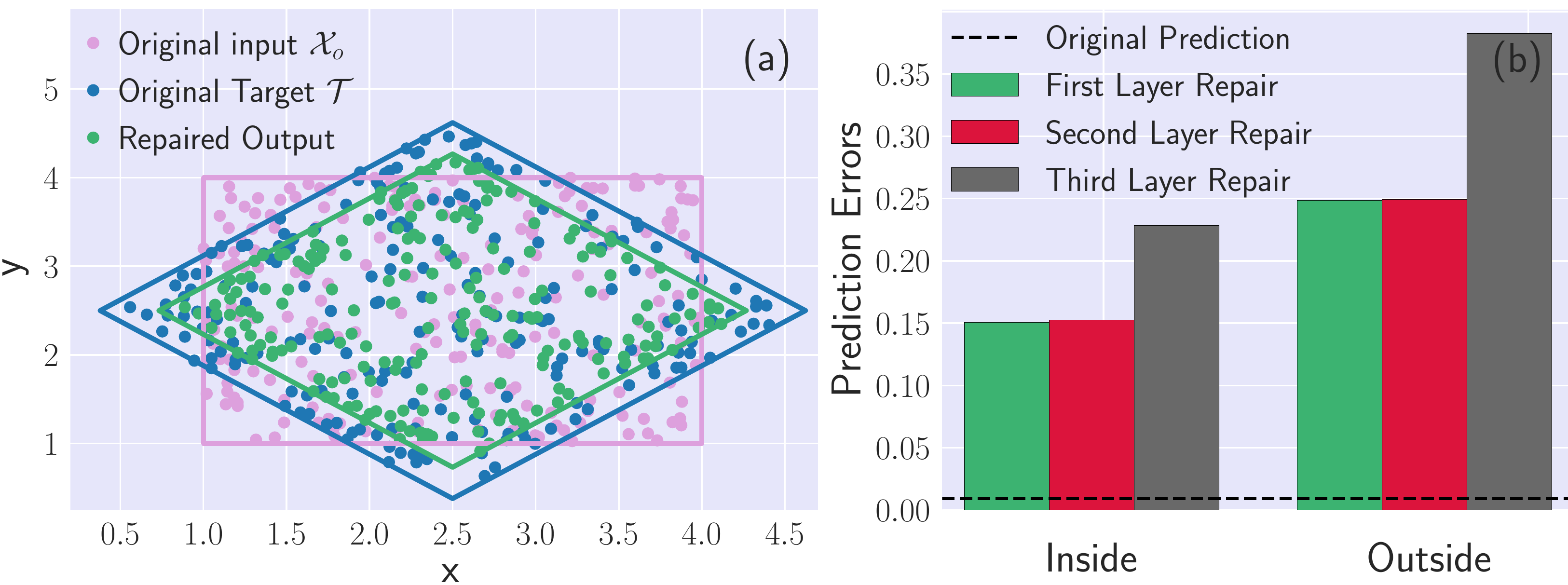}
    \caption{In-place rotation: (a) predictions after applying repair to the first layer, (b) Mean Square Error (MSE) between testing and original targets.}
    \label{fig: affine-trans}
\end{figure}

Figure \ref{fig: affine-trans} (a) demonstrates network predictions after applying repair to the first layer $\mathbf{x}^0\xrightarrow{\mathbf{W}^1,\mathbf{b}^1}\mathbf{x}^1$. 
As it is shown, the predicate $\Psi(\mathbf{y},\mathbf{x}^0)$ pushes the predicted outputs inside the green set. 
The bar plot in Fig. \ref{fig: affine-trans} (b) illustrates the Mean Square Error (MSE) of the network predictions after applying repair to different layers (showed with colored bars) compared to the original network predictions MSE (showed with dashed line). 
This plot separates the prediction errors of samples based on their targets whether located inside or outside of the green set. 
Clearly, prediction error for the points with targets inside the green set is lower than the points with targets outside the green set. 
The prediction performance improves by moving from the last layer to the first layer for both inside and outside cases. The large error reduction by hidden layers is due to the capture of non-linearities in their forward pass. However, the last layer only has the scaling factor, thereby, results smaller error reduction compared to the hidden layers.
Moreover, to achieve a significant error reduction, we can select any hidden layer since moving among hidden layers does not result a large error change. 
In practice, we observed that repairing the latter hidden layers helps Gurobi solver to find a solution faster since the number of integer variables created by modeling ReLU activation function is increased by $2^l$ moving backward from each hidden layer $l$. 

In this experiment, the optimal weight deviations for each layer are $\delta_1=0.391$, $\delta_2 = 0.4408$, and $\delta_3 = 0.7605$ for the first, second, and third layers, respectively. It implies that repairing by moving backward from the last layer needs smaller weight deviation to adjust the network's behavior.

\subsection{Forward Kinematics}
\begin{figure}[t]
    \centering
    \includegraphics[scale=0.28]{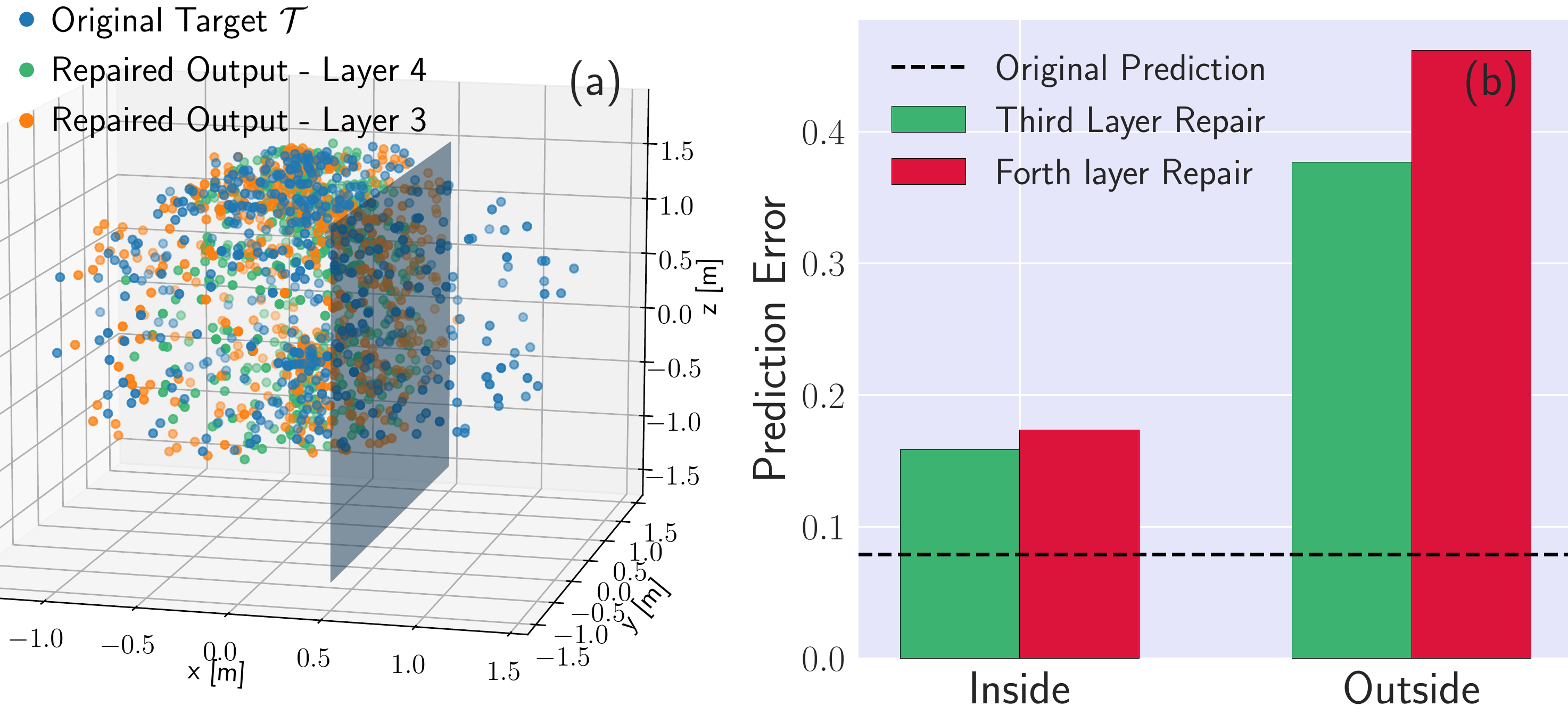}
    \caption{Forward kinematics: (a) output predictions, (b) Mean Square Error (MSE) between testing and original targets.}
    \label{fig: forward-kin}
\end{figure}

In our second test, we bound the output of a three-hidden-layer network $N\!N_o$ ($L=3$) with $20$ nodes in each layer that learned the forward kinematics of a 6-DOF KUKA robot arm \cite{kazi2005research}. 
The forward kinematics function $\textsc{Fk}\!:\mathcal{X}_0\rightarrow\mathcal{T}$ for this robot is a mapping from the $6$ joint angles to the location of end-effector in 3D space. 
To train the original network, we generate a forward kinematics data set using RoboAnalyzer simulator \cite{gupta2017roboanalyzer}. 
Here, the inputs to the $N\!N_o$ are the $\sin(\theta_q)$ and $\cos(\theta_q)$ of each joint angle $\theta_q$ for $\{q\}^6_{q=1}$ ($12$ inputs in total). 
It helps to remove the discontinuity that occurs by mapping $\theta$ to $[-\pi,\pi]$. 
Given the trigonometrically filtered input angles, we define the original input space as $\mathcal{X}_o = \big\{\mathbf{x}^0\in \mathbb{R}^{12}\;|\; -1\leq x^0_j\leq1 \text{ for }\{j\}^{12}_{j=1}\big\}$. 
The output space $\mathcal{T}$ is then defined as $\mathcal{T}=\{\mathbf{y}\in\mathbb{R}^{3}\;|\;\mathbf{y}=\textsc{Fk}(\mathbf{x}_0)\text{ for }\mathbf{x}_0\in\mathcal{X}_o\}$, where the outputs $\mathbf{y}=[x,y,z]$ are the coordinates of end-effector in 3D space. 
We repair the network to satisfy a predicate defined as $\Psi(\mathbf{y},\mathbf{x}^0)=\{\mathbf{y}\in\mathcal{T}\;|\; x\leq 0.5\}$ for $\mathcal{X}_r=\mathcal{X}_o$.
Here, $x>5$ can be the space of a human worker who is jointly interacting with the robot. 
So, $\Psi(\mathbf{y},\mathbf{x}^0)$ bounds the location of the end-effector to not operate in this space. We repair the network with $N = 500$ samples.  
Figure \ref{fig: forward-kin} illustrates the results of applying network repairs on the forth and third layers. 
As shown in Fig. \ref{fig: forward-kin} (a), the repaired network successfully pushes the outputs inside the robot's operation region and has an accurate prediction performance for the points inside this bound (see Fig. \ref{fig: forward-kin} (b)). 
Prediction errors also demonstrate that the repair of the third layer has lower error compared to the forth layer. 
In this experiment, the optimal weight deviations for each layer are $\delta_3 = 0.5761$, and $\delta_4 = 0.8024$ for the third and forth layers, respectively.

\subsection{Unmanned Aircraft Collision Avoidance system X (ACAS Xu)} 
\begin{figure}[t]
    \centering
    \includegraphics[scale=0.34]{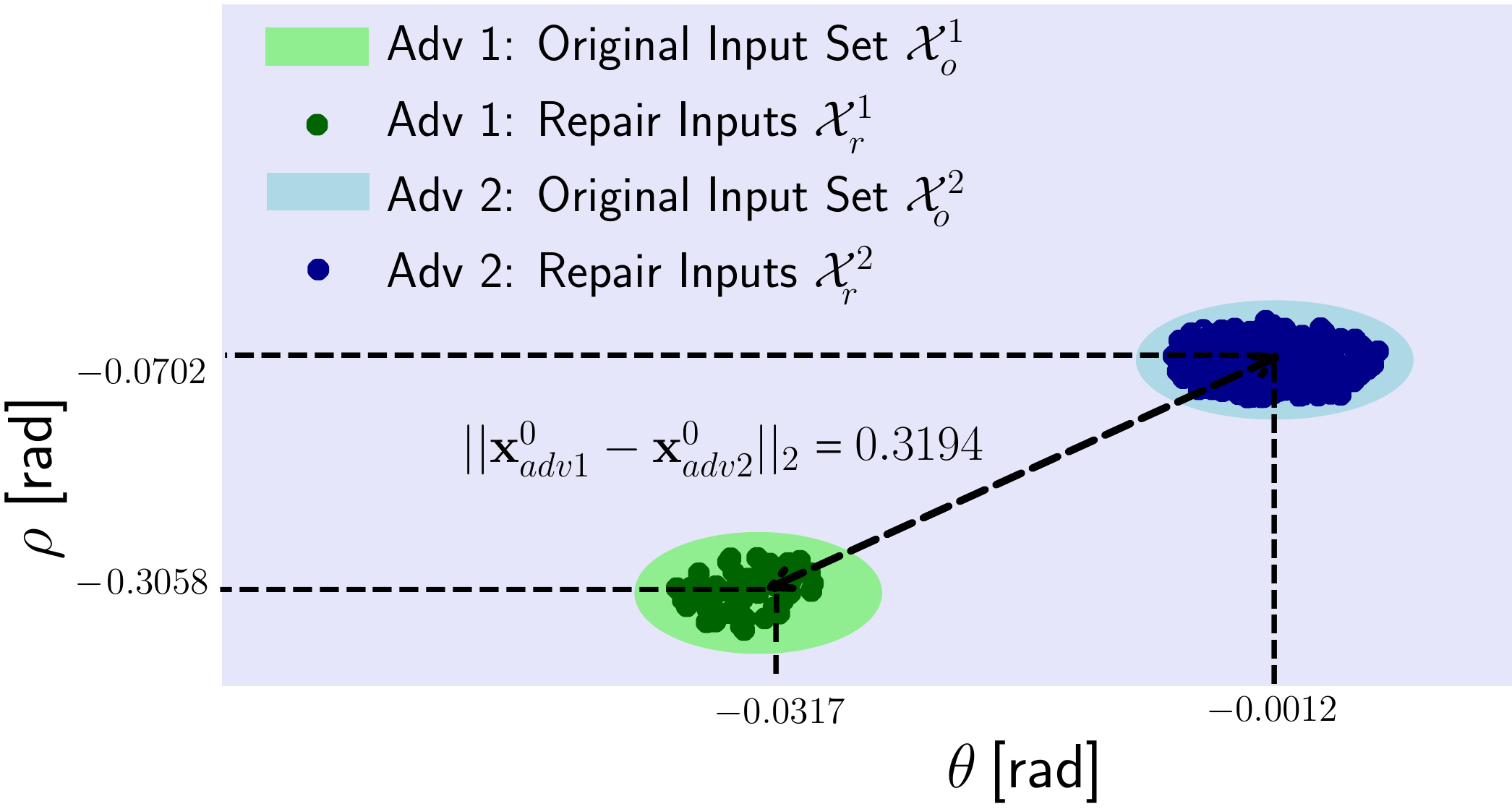}
    \caption{Collision Avoidance System: projection of repaired regions on $\theta$ and $\rho$ input spaces shown in green and blue. Misclassified points shown with dots. $\lVert \mathbf{x}^0_{adv1}-\mathbf{x}^0_{adv2}\rVert_2$ represents the average Euclidean distance between the two local repaired regions.}
    \label{fig: classification}
\end{figure}

ACAS Xu is a variant of Aircraft Collision Avoidance System X (ACAS X) that provides horizontal manoeuvre advisories for unmanned aircraft \cite{kochenderfer2011robust}. This system uses a look up table that produces 5 different horizontal advisory outputs of clear-of-conflict ($COC$), weak right/left heading rates of $\pm 1.5^{\circ} / s$ ($WR$ and $WL$), and strong right/left heading rates of $\pm 3^{\circ} / s$ degrees per second ($SR$ and $SL$), i.e. $\mathbf{y}=[COC,WR,WL,SR,SL]$. The inputs of system are 7 sensor measurements of distance from ownship to intruder $\rho$ ($m$), angle to intruder relative to ownership heading
direction $\theta$ ($rad$), heading angle of intruder relative to ownship
heading direction $\psi$ ($rad$), speed of ownship $v_{own}$ ($m/s$), speed of intruder $v_{int}$ ($m/s$), time until loss of vertical separation $\tau$ ($s$), and previous advisory $a_{prev}$ ($^{\circ} / s$). The advisory in $\mathbf{y}$ with the minimum value is the selected advisory for a given input. To improve storage efficiency, an array of 45 Deep Neural Networks (DNN) are trained in \cite{julian2016policy} for the discretized combinations of $\tau$ and $a_{prev}$ to learn the look up table. Inputs of each DNN are $\mathbf{x}^0=[\rho,\theta,\psi,v_{own},v_{int}]$. Each DNN has 6 hidden layers with 50 ReLU activation nodes in each layer. DNNs return a horizontal advisory given the input sensor measurements. Given the application of this system, it is very important for the system to return correct advisories. In this experiment, we focus on the DNN trained for $a_{prev} = WL$ and $\tau = 100$. Following the look up table, the desired output property of this network is to return $WL$ or $COC$ horizontal advisories for the input space \begin{align*}
    \mathcal{X}_o=\{\mathbf{x}^0=&[\rho,\theta,\psi,v_{own},v_{int}]\in\mathbb{R}^5\;|\;-0.33\leq\rho\leq 0.68 \\
    &\text{ and } -0.5\leq \theta,\psi,v_{own},v_{int}\leq 0.5\}.
\end{align*}
Thus, the predicate is defined as $\Psi(\mathbf{y},\mathbf{x}^0)=\{\mathbf{y}\in\mathbb{R}^5\;|\;COC,WL<WR,SR,SL\}.$ To find input samples in $\mathcal{X}_o$ that violate $\Psi(\mathbf{y},\mathbf{x}^0)$, we use Marabou verification tool proposed in \cite{katz2019marabou}. 
Figure \ref{fig: classification} illustrates the projection of local misclassified set of inputs $\mathcal{X}^1_r$ and $\mathcal{X}^2_r$, shown with green and blue dots respectively, on $\theta$ and $\rho$ input spaces. 
Our method successfully corrects the local misclassified set of points $\mathcal{X}^1_r\subseteq\mathcal{X}^1_o$ and $\mathcal{X}^2_r\subseteq\mathcal{X}^2_o$ by applying our NN repair technique on the last layer. 
In this experiment, $\delta=9.8\times10^{-5}$ and $\delta=1.14\times10^{-3}$ for repairing $\mathcal{X}^1_r$ and $\mathcal{X}^2_r$, respectively, that demonstrates the satisfaction of property $\Psi(\mathbf{y},\mathbf{x}^0)$ with just a small deviation of last layer's original weights.

\subsection{Safe Controller}
\begin{figure}[t]
    \centering
    \begin{subfigure}[t]{0.23\textwidth}
        \centering
        \includegraphics[scale=0.233]{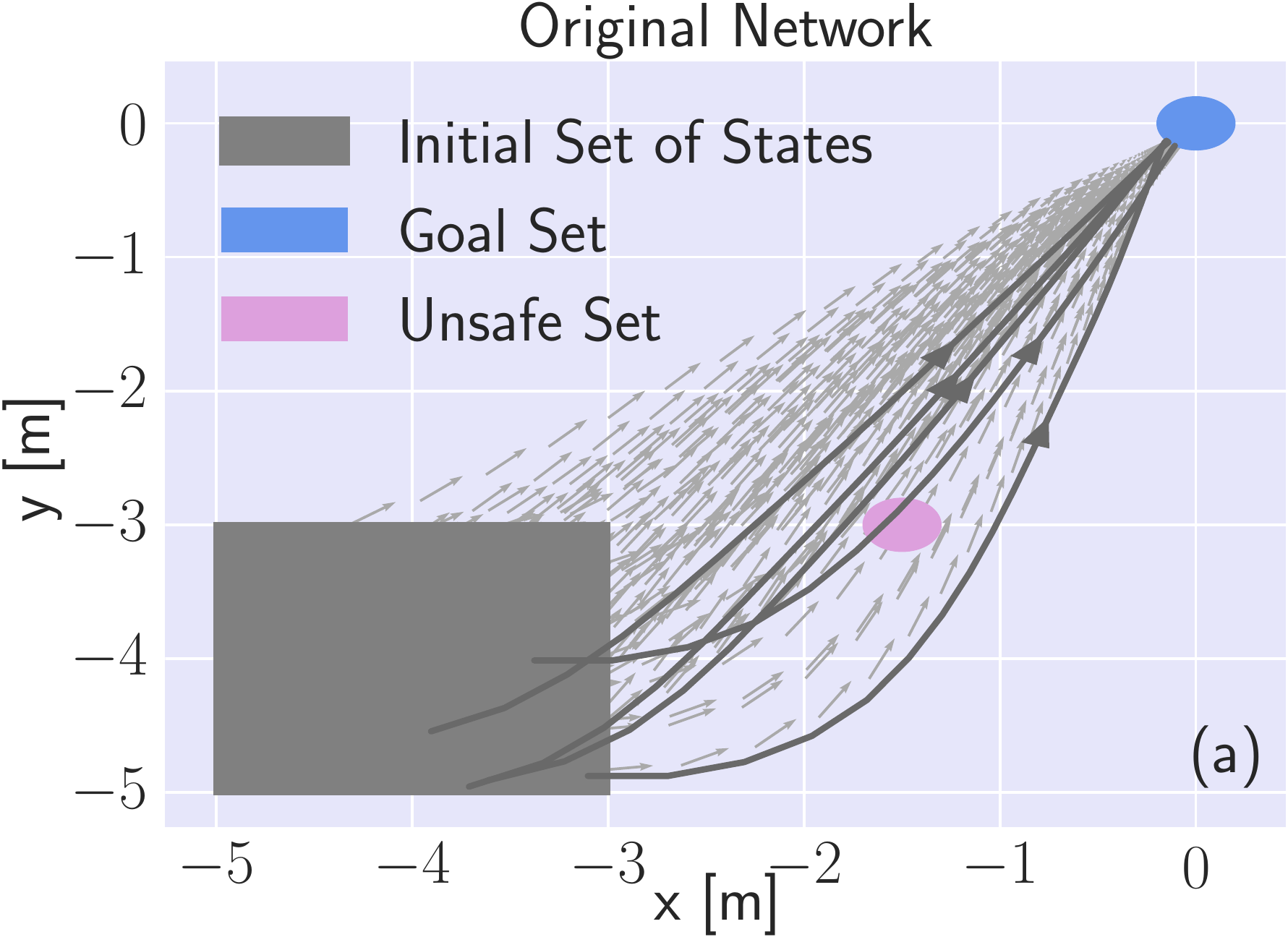}
    \end{subfigure}%
    \hfill
    \begin{subfigure}[t]{0.24\textwidth}
        \centering
        \includegraphics[scale=0.233]{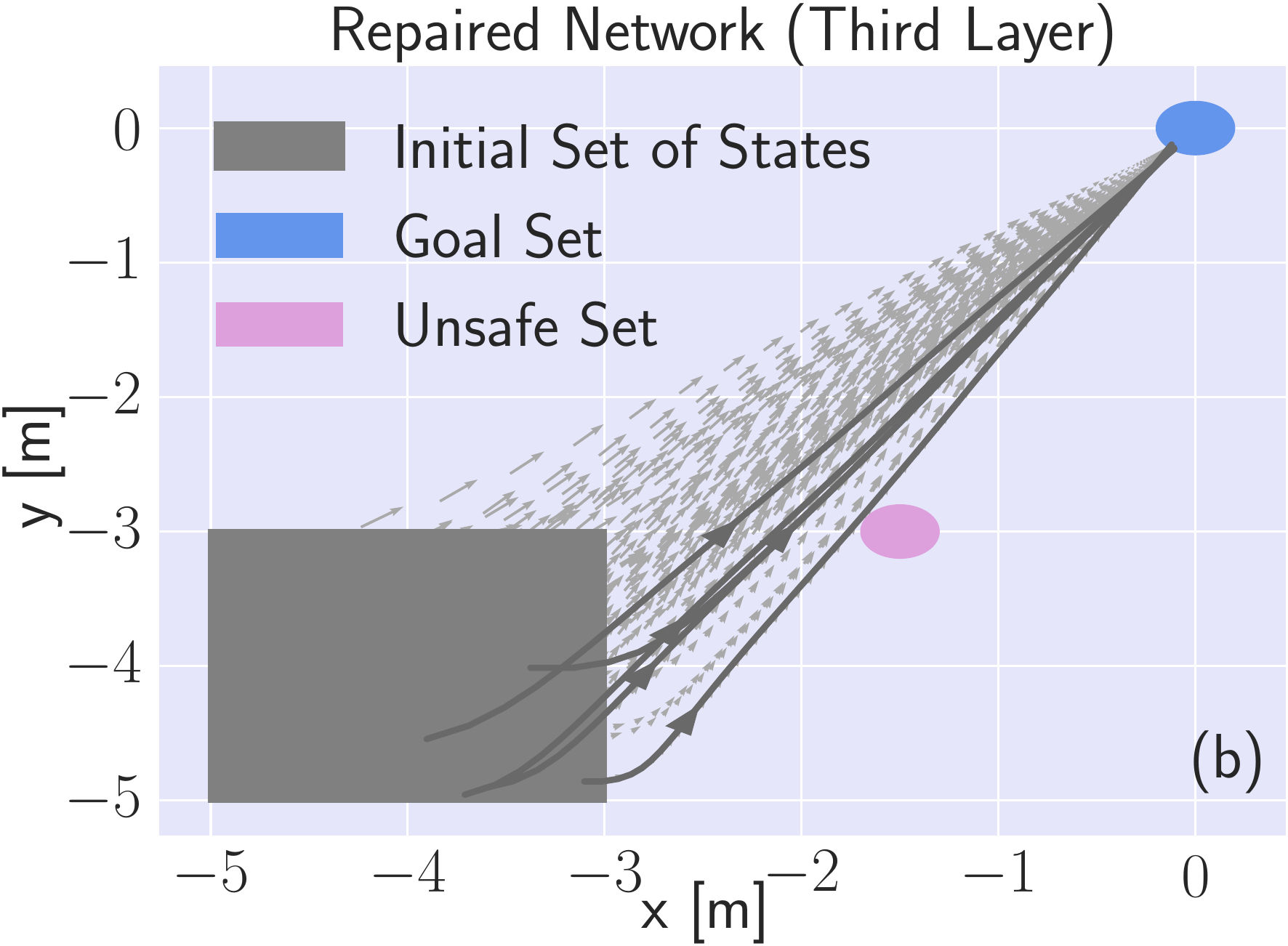}
    \end{subfigure}

    \caption{Safe controller: (a) simulated trajectories of system under learned policy by the original network, (b) simulated trajectories of system under repaired policy.}
    \label{fig:controller}
\end{figure}

In our last experiment, we repair a mobile robot point-to-goal controller to avoid a static obstacle. 
Consider a robot that is following a unicycle model as 
\begin{align}\label{eq: robot-model}
    \dot{\mathbf{x}}=\mathbf{g}(\mathbf{x})\mathbf{u} ={\small\begin{bmatrix}
    \cos(\theta) & 0\\
    \sin(\theta) & 0\\
     0 & 1
    \end{bmatrix}}\mathbf{u}.
\end{align}
where states are $\mathbf{x}\!=\![x,y,\theta ]^T\in \mathcal{X}\!\in\! \mathbb{R}^2\!\times\![-\pi,\pi)$ and control inputs are $\mathbf{u}=[v,\omega]^T\in\mathcal{U}\in \mathbb{R}^2$.
The variables $x$, $y$, $\theta$ denote the longitudinal and lateral positions of the robot and heading angle, respectively. 
The controls $v$ and $\omega$ represent the linear and angular velocities of robot, respectively. 
We first use imitation learning to train a neural network controller $N\!N_o$ that imitates a QP-based Controller $\textsc{Ctr}:\mathcal{X}\rightarrow\mathcal{U}$ as an expert.
$N\!N_o$ receives the state of the robot as the input $\mathbf{x}^0=\mathbf{x}$ ($\mathcal{X}_o=\mathcal{X}$) and outputs the control action, i.e. $\mathbf{y}=\mathbf{u}$ ($\mathcal{T}=\mathcal{U}$). 
The training samples are collected from $\mathcal{X}_o\times\mathcal{T}=\big\{\{(\mathbf{x}^0_n,\mathbf{t}_n)\}^N_{n=1}\in \mathcal{X}\!\times\!\mathcal{U}\;|\; \mathbf{x}^0_1\in\mathcal{X}_{init},\mathbf{x}^0_N\in\mathcal{X}_g,\mathbf{t}_n=\textsc{Ctr}(\mathbf{x}_n^0)\big\}$.
This controller steers the robot from an initial set of states $\mathcal{X}_{init}=\{\mathbf{x}^0\in \mathcal{X}_o ~\lvert ~x,y\in[-5,-3],\theta\in[0,\frac{\pi}{2}]\}$ to the goal set $\mathcal{X}_g=\{\mathbf{x}^0\in\mathcal{X}_o\;|\;\big\lVert V(\mathbf{x}^0) \leq 0 \}$. 
Here $V:\mathcal{X}_o\rightarrow\mathbb{R}$ is a continuously differentiable function  
\begin{align}\label{eq: lyap}
    V(\mathbf{x}^0)=\big\lVert[x,y]^T-[0,0]\big\rVert^2_2 - 0.2^2.
\end{align}
We model $N\!N_o$ to have 2 hidden layers with 10 ReLU nodes per layer. 
Figure \ref{fig:controller} (a) shows the trajectories of system under QP-based learned policy. 
Readers are referred to \cite{yaghoubi2020training} for more details on using QP-based controllers in imitation learning. 

Now assume for the same control task, the robot needs to avoid an unsafe region (the ellipsoidal violet region in Fig. \ref{fig:controller}) defined as 
\begin{align}\label{eq: unsafe set}
    \mathcal{X}_{unsafe} = \{ \mathbf{x}^0 \in \mathcal{X}_o ~\lvert ~h(\mathbf{x}^0)< 0\},
\end{align}
where $h:\mathcal{X}_o\rightarrow\mathbb{R}$ is a continuously differentiable safety measure that represents a static obstacle defined as
\begin{align}\label{eq: cbf}
    h(\mathbf{x}^0) = \lVert[x,y]^T - [-1.5,-3]^T\rVert_2^2 - 0.2^2.
\end{align}
We treat $h(\mathbf{x}^0)$ as a Control Barrier Function (CBF) \cite{ames2019control} to derive a safety predicate for the output $\mathbf{y}=\mathbf{u}$ of $N\!N_o$ controller under which the states $\mathbf{x}$ of system ($\ref{eq: robot-model}$) that start in $\mathcal{X}\setminus \mathcal{X}_{unsafe}$ never enters the unsafe set $\mathcal{X}_{unsafe}$.
\begin{definition}[Control Barrier Function (CBF)]
Given the nonlinear control affine system (\ref{eq: robot-model}) and an unsafe set $\mathcal{X}_{unsafe}$, a continuously differentiable function $h:\mathcal{X}\rightarrow\mathbb{R}$ is a CBF for this system if there exists an extended $\mathcal{K}$ class function $\alpha:\mathbb{R}\rightarrow\mathbb{R}$ (strictly increasing and $\alpha(0)=0$) such that
\begin{align}
    &h(\mathbf{x})\geq 0 &&\forall \mathbf{x}\in \mathcal{X}\setminus \mathcal{X}_{unsafe},\label{eq: CBF1}\\
    &h(\mathbf{x})< 0 &&\forall \mathbf{x}\in \mathcal{X}_{unsafe},\label{eq: CBF2}\\
    &L_g h(\mathbf{x}) \mathbf{u} +\alpha(h(\mathbf{x}))\!\geq\! 0 &&\forall \mathbf{x}\in \mathcal{X}\setminus \mathcal{X}_{unsafe}, \label{eq: CBF3}
\end{align}
where $L_g h(\mathbf{x})$ is the first order Lie derivative of $\frac{\partial h}{\partial \mathbf{x}}^T\mathbf{g}$.
\end{definition}
If the control $\mathbf{u}$ satisfies (\ref{eq: CBF3}), then the state of system (\ref{eq: robot-model}) never leaves $\mathcal{X}\setminus \mathcal{X}_{unsafe}$.
Our defined candidate function (\ref{eq: cbf}) satisfies the conditions (\ref{eq: CBF1}) and (\ref{eq: CBF2}) given the defined  unsafe set (\ref{eq: unsafe set}). 
Now to satisfy (\ref{eq: CBF3}), for all $\mathbf{x}^0\!\in\!\mathcal{X}_r=\mathcal{X}_o\setminus\mathcal{X}_{unsafe}$ we impose the predicate $\Psi_1(\mathbf{x}^0,\mathbf{y})=\{\mathbf{y}\!\in \!\mathcal{U}\;|\;L_g h(\mathbf{x}^0) \mathbf{y} +\gamma h(\mathbf{x}^0)\!\geq\! 0\}$ to the output $\mathbf{y}$ of $N\!N_o$, where $\gamma \geq 0$.
To ensure that the repaired controller steers the robot toward the goal set $\mathcal{X}_g$, we also treat $V(\mathbf{x}^0)$, defined in (\ref{eq: lyap}), as a Lyapunov-like function. 
\begin{definition}[Lyapunov-like Function (CLF)]
Given the nonlinear control affine system (\ref{eq: robot-model}) and a goal set $\mathcal{X}_g$, a continuously differentiable function $V:\mathcal{X}\rightarrow\mathbb{R}$ is a CLF for this system if
\begin{align}
    &V(\mathbf{x})> 0 &&\forall \mathbf{x}\in \mathcal{X}\setminus \mathcal{X}_g,\label{eq: CLF1}\\
    &V(\mathbf{x})\leq 0 &&\forall \mathbf{x}\in \mathcal{X}_g,\label{eq: CLF2}\\
    &L_g V(\mathbf{x}) \mathbf{u} \!\leq\! 0 &&\forall \mathbf{x}\in \mathcal{X}\setminus \mathcal{X}_g. \label{eq: CLF3}
\end{align}
\end{definition}
If there exists a control $\mathbf{u}$ that satisfies (\ref{eq: CLF3}), then the states of system (\ref{eq: robot-model}) reach the goal set $\mathcal{X}_g$ where $V(\mathbf{x})\leq 0$. 
Similar to the safety condition, $V(\mathbf{x}^0)$, defined in (\ref{eq: lyap}), satisfies the conditions (\ref{eq: CLF1}) and (\ref{eq: CLF2}). So for all $\mathbf{x}^0\!\in\!\mathcal{X}_r$ we impose  
$\Psi_2(\mathbf{x}^0,\mathbf{y})=\{\mathbf{y}\in \mathcal{U}\;|\;L_g V(\mathbf{x}^0) \mathbf{y} \leq \beta\}$ to the output $\mathbf{y}$ of $N\!N_o$ to satisfy condition (\ref{eq: CLF3}).
Here $\beta\geq 0$ is a slack variable that minimally relaxes the predicate $\Psi_2$ when enforcing the goal-reaching constraint to $\mathbf{u}$ contradicts the safety condition $\Psi_1$.
For further details on Lyapunov-like functions, CBFs, and their application in safe mobile robot navigation, readers are referred to \cite{ames2019control,yaghoubi2020risk,majd2021safe}. 

Given the linear inequality constraints $\Psi_1$ and $\Psi_2$, we repair the last layer of $N\!N_o$ with 100 reference trajectories. 
As illustrated in Fig. \ref{fig:controller} (b), the generated trajectories of system (\ref{eq: robot-model}) by the repaired policy successfully avoid the unsafe region showed in violet. 
The repaired controller decreases the linear and angular velocities $v$ and $\omega$ for the states close to the unsafe region while the repaired policy converges to the original policy when states get farther from the unsafe region.
In this experiment, the optimal weight deviation is $\delta=2.45$.

\section{Conclusion and Future Work}\vspace{-3pt}
In this paper, we provided a framework for minimal layer-wise neural network repair. 
For a repair, we adjust a trained neural network to satisfy a set of predicates that impose constraints on the output of NN for a given set of inputs of interest. 
The experimental results demonstrated the success of our framework in repairing the trained networks to satisfy a set of predicates while maintaining the performance of the original network. 

Inspired by our theoretical and experimental results, one is able to improve, change, and guarantee different aspects of performance of a trained neural network. 
While we showed the capability in our framework in addressing safety, model improvement, and verification problems in different robot learning applications, MIQP formulation of network repair is not an easy problem to solve for high-dimensional robotic systems with large input spaces. 
Our future work aims to address this problem by relaxing or decreasing the number of integer variables of our formulation. 
Additionally, we aim to investigate an efficient sampling mechanism for the network repair that can help to decrease the number of integer variables as well. 
In our experiments, we also observed that repairing the last layer did not necessarily satisfy a given property and, in some cases, it caused infeasibility. 
While we did not observe this issue in repairing the hidden layers, we aim to explore more cases where infeasibility occurs. 
Finally, we plan to create a public available neural network repair Python package integrated with TensorFlow model architectures \cite{tensorflow2015-whitepaper} in our future work.


\bibliographystyle{IEEEtran}
\bibliography{main.bbl}

\begin{thebibliography}{10}
\providecommand{\url}[1]{#1}
\csname url@samestyle\endcsname
\providecommand{\newblock}{\relax}
\providecommand{\bibinfo}[2]{#2}
\providecommand{\BIBentrySTDinterwordspacing}{\spaceskip=0pt\relax}
\providecommand{\BIBentryALTinterwordstretchfactor}{4}
\providecommand{\BIBentryALTinterwordspacing}{\spaceskip=\fontdimen2\font plus
\BIBentryALTinterwordstretchfactor\fontdimen3\font minus
  \fontdimen4\font\relax}
\providecommand{\BIBforeignlanguage}[2]{{%
\expandafter\ifx\csname l@#1\endcsname\relax
\typeout{** WARNING: IEEEtran.bst: No hyphenation pattern has been}%
\typeout{** loaded for the language `#1'. Using the pattern for}%
\typeout{** the default language instead.}%
\else
\language=\csname l@#1\endcsname
\fi
#2}}
\providecommand{\BIBdecl}{\relax}
\BIBdecl

\bibitem{RiviereEtAl2020ral}
B.~Riviere, W.~Honig, Y.~Yue, and S.-J. Chung, ``Glas: Global-to-local safe
  autonomy synthesis for multi-robot motion planning with end-to-end
  learning,'' vol.~5, no.~3, pp. 4249--4256.

\bibitem{ZhouEtAl2019iros}
S.~Zhou, M.~J. Phielipp, J.~A. Sefair, S.~I. Walker, and H.~B. Amor, ``Clone
  swarms: Learning to predict and control multi-robot systems by imitation,''
  in \emph{IEEE/RSJ International Conference on Intelligent Robots and Systems
  (IROS)}.

\bibitem{FanLLP2020ijrr}
T.~Fan, P.~Long, W.~Liu, and J.~Pan, ``Distributed multi-robot collision
  avoidance via deep reinforcement learning for navigation in complex
  scenarios,'' vol.~39, no.~7, pp. 856--892.

\bibitem{JulianEtAl2016dasc}
K.~D. Julian, J.~Lopez, J.~S. Brush, M.~P. Owen, and M.~J. Kochenderfer,
  ``Policy compression for aircraft collision avoidance systems,'' in
  \emph{IEEE/AIAA 35th Digital Avionics Systems Conference (DASC)}.

\bibitem{PanEtAl2018rss}
Y.~Pan, C.-A. Cheng, K.~Saigol, K.~Lee, X.~Yan, E.~Theodorou, and B.~Boots,
  ``Agile autonomous driving using end-to-end deep imitation learning,'' in
  \emph{Robotics: Science and Systems}.

\bibitem{KuuttiEtAl2021its}
S.~Kuutti, R.~Bowden, Y.~Jin, P.~Barber, and S.~Fallah, ``A survey of deep
  learning applications to autonomous vehicle control,'' vol.~22, no.~2, pp.
  712--733.

\bibitem{StricklandFBA2018icra}
M.~Strickland, G.~Fainekos, and H.~B. Amor, ``Deep predictive models for
  collision risk assessment in autonomous driving,'' in \emph{IEEE
  International Conference on Robotics and Automation (ICRA)}, 2018.

\bibitem{TianEtAl2018icse}
Y.~Tian, K.~Pei, S.~Jana, and B.~Ray, ``Deeptest: Automated testing of
  deep-neural-network-driven autonomous cars,'' in \emph{40th International
  Conference on Software Engineering ({ICSE})}, 2018.

\bibitem{Dreossi2019}
T.~Dreossi, A.~Donze, and S.~A. Seshia, ``Compositional falsification of
  cyber-physical systems with machine learning components,'' vol.~63, p.
  1031–1053, 2019.

\bibitem{TuncaliEtAl2020tiv}
C.~E. Tuncali, G.~Fainekos, D.~Prokhorov, H.~Ito, and J.~Kapinski,
  ``Requirements-driven test generation for autonomous vehicles with machine
  learning components,'' \emph{IEEE Transactions on Intelligent Vehicles},
  vol.~5, pp. 265--280, 2020.

\bibitem{SunKS2019hscc}
X.~Sun, H.~Khedr, and Y.~Shoukry, ``Formal verification of neural network
  controlled autonomous systems,'' in \emph{22nd ACM International Conference
  on Hybrid Systems: Computation and Control}, 2019.

\bibitem{IvanovEtAl2020}
R.~Ivanov, T.~J. Carpenter, J.~Weimer, R.~Alur, G.~J. Pappas, and I.~Lee,
  ``Verifying the safety of autonomous systems with neural network
  controllers,'' vol.~20, no.~1.

\bibitem{DuttaEtAl2018adhs}
S.~Dutta, S.~Jha, S.~Sankaranarayanan, and A.~Tiwari, ``Learning and
  verification of feedback control systems using feedforward neural networks,''
  in \emph{Analysis and Design of Hybrid Systems}, 2018.

\bibitem{TjengXT2019iclr}
V.~Tjeng, K.~Y. Xiao, and R.~Tedrake, ``Evaluating robustness of neural
  networks with mixed integer programming,'' in \emph{7th International
  Conference on Learning Representations {(ICLR)}}.

\bibitem{YaghoubiF2019tecs}
S.~Yaghoubi and G.~Fainekos, ``Worst-case satisfaction of stl specifications
  using feedforward neural network controllers: A lagrange multipliers
  approach,'' \emph{ACM Transactions on Embedded Computing Systems}, vol.~18,
  no.~5S, 2019.

\bibitem{DreossiJS2018arxiv}
T.~Dreossi, S.~Jha, and S.~A. Seshia, ``Semantic adversarial deep learning,''
  \emph{arXiv:1804.07045v2}, 2018.

\bibitem{DreossiGYKSS2018ijcai}
T.~Dreossi, S.~Ghosh, X.~Yue, K.~Keutzer, A.~L. Sangiovanni{-}Vincentelli, and
  S.~A. Seshia, ``Counterexample-guided data augmentation,'' in
  \emph{Proceedings of the Twenty-Seventh International Joint Conference on
  Artificial Intelligence, {IJCAI}}, 2018, pp. 2071--2078.

\bibitem{goldberger2020minimal}
B.~Goldberger, G.~Katz, Y.~Adi, and J.~Keshet, ``Minimal modifications of deep
  neural networks using verification.'' in \emph{LPAR}, vol. 2020, 2020, p.
  23rd.

\bibitem{DongEtAl2021arxiv}
G.~Dong, J.~Sun, J.~Wang, X.~Wang, T.~Dai, and X.~Wang, ``Towards repairing
  neural networks correctly.''

\bibitem{CruzFS2021arxiv}
U.~S. Cruz, J.~Ferlez, and Y.~Shoukry, ``Safe-by-repair: A convex optimization
  approach for repairing unsafetwo-level lattice neural network controllers.''

\bibitem{balas1979disjunctive}
E.~Balas, ``Disjunctive programming,'' \emph{Annals of discrete mathematics},
  vol.~5, pp. 3--51, 1979.

\bibitem{tsay2021partition}
C.~Tsay, J.~Kronqvist, A.~Thebelt, and R.~Misener, ``Partition-based
  formulations for mixed-integer optimization of trained relu neural
  networks,'' \emph{arXiv preprint arXiv:2102.04373}, 2021.

\bibitem{belotti2011disjunctive}
P.~Belotti, L.~Liberti, A.~Lodi, G.~Nannicini, A.~Tramontani \emph{et~al.},
  ``Disjunctive inequalities: applications and extensions,'' \emph{Wiley
  Encyclopedia of Operations Research and Management Science}, vol.~2, pp.
  1441--1450, 2011.

\bibitem{hart2011pyomo}
W.~E. Hart, J.-P. Watson, and D.~L. Woodruff, ``Pyomo: modeling and solving
  mathematical programs in python,'' \emph{Mathematical Programming
  Computation}, vol.~3, no.~3, pp. 219--260, 2011.

\bibitem{bynum2021pyomo}
M.~L. Bynum, G.~A. Hackebeil, W.~E. Hart, C.~D. Laird, B.~L. Nicholson, J.~D.
  Siirola, J.-P. Watson, and D.~L. Woodruff, \emph{Pyomo--optimization modeling
  in python}, 3rd~ed.\hskip 1em plus 0.5em minus 0.4em\relax Springer Science
  \& Business Media, 2021, vol.~67.

\bibitem{chen2018pyomo}
Q.~Chen, E.~S. Johnson, J.~D. Siirola, and I.~E. Grossmann, ``Pyomo. gdp:
  Disjunctive models in python,'' in \emph{Computer Aided Chemical
  Engineering}.\hskip 1em plus 0.5em minus 0.4em\relax Elsevier, 2018, vol.~44,
  pp. 889--894.

\bibitem{gurobi}
\BIBentryALTinterwordspacing
{Gurobi Optimization, LLC}, ``{Gurobi Optimizer Reference Manual},'' 2021.
  [Online]. Available: \url{https://www.gurobi.com}
\BIBentrySTDinterwordspacing

\bibitem{kazi2005research}
A.~Kazi and R.~Bischoff, ``From research to products: the kuka perspective on
  european research projects,'' \emph{IEEE robotics \& automation magazine},
  vol.~12, no.~3, pp. 78--84, 2005.

\bibitem{gupta2017roboanalyzer}
V.~Gupta, R.~G. Chittawadigi, and S.~K. Saha, ``Roboanalyzer: robot
  visualization software for robot technicians,'' in \emph{Proceedings of the
  Advances in Robotics}, 2017, pp. 1--5.

\bibitem{kochenderfer2011robust}
M.~J. Kochenderfer and J.~Chryssanthacopoulos, ``Robust airborne collision
  avoidance through dynamic programming,'' \emph{Massachusetts Institute of
  Technology, Lincoln Laboratory, Project Report ATC-371}, vol. 130, 2011.

\bibitem{julian2016policy}
K.~D. Julian, J.~Lopez, J.~S. Brush, M.~P. Owen, and M.~J. Kochenderfer,
  ``Policy compression for aircraft collision avoidance systems,'' in
  \emph{2016 IEEE/AIAA 35th Digital Avionics Systems Conference (DASC)}.\hskip
  1em plus 0.5em minus 0.4em\relax IEEE, 2016, pp. 1--10.

\bibitem{katz2019marabou}
G.~Katz, D.~A. Huang, D.~Ibeling, K.~Julian, C.~Lazarus, R.~Lim, P.~Shah,
  S.~Thakoor, H.~Wu, A.~Zelji{\'c} \emph{et~al.}, ``The marabou framework for
  verification and analysis of deep neural networks,'' in \emph{International
  Conference on Computer Aided Verification}.\hskip 1em plus 0.5em minus
  0.4em\relax Springer, 2019, pp. 443--452.

\bibitem{yaghoubi2020training}
S.~Yaghoubi, G.~Fainekos, and S.~Sankaranarayanan, ``Training neural network
  controllers using control barrier functions in the presence of
  disturbances,'' in \emph{2020 IEEE 23rd International Conference on
  Intelligent Transportation Systems (ITSC)}.\hskip 1em plus 0.5em minus
  0.4em\relax IEEE, 2020, pp. 1--6.

\bibitem{ames2019control}
A.~D. Ames, S.~Coogan, M.~Egerstedt, G.~Notomista, K.~Sreenath, and P.~Tabuada,
  ``Control barrier functions: Theory and applications,'' in \emph{2019 18th
  European Control Conference (ECC)}.\hskip 1em plus 0.5em minus 0.4em\relax
  IEEE, 2019, pp. 3420--3431.

\bibitem{yaghoubi2020risk}
S.~Yaghoubi, K.~Majd, G.~Fainekos, T.~Yamaguchi, D.~Prokhorov, and B.~Hoxha,
  ``Risk-bounded control using stochastic barrier functions,'' \emph{IEEE
  Control Systems Letters}, vol.~5, no.~5, pp. 1831--1836, 2020.

\bibitem{majd2021safe}
K.~Majd, S.~Yaghoubi, T.~Yamaguchi, B.~Hoxha, D.~Prokhorov, and G.~Fainekos,
  ``Safe navigation in human occupied environments using sampling and control
  barrier functions,'' in \emph{IEEE/RSJ International Conference on
  Intelligent Robots and Systems (IROS)}, 2021.

\bibitem{tensorflow2015-whitepaper}
\BIBentryALTinterwordspacing
M.~Abadi, A.~Agarwal, P.~Barham, E.~Brevdo, Z.~Chen, C.~Citro, G.~S. Corrado,
  A.~Davis, J.~Dean, M.~Devin, S.~Ghemawat, I.~Goodfellow, A.~Harp, G.~Irving,
  M.~Isard, Y.~Jia, R.~Jozefowicz, L.~Kaiser, M.~Kudlur, J.~Levenberg,
  D.~Man\'{e}, R.~Monga, S.~Moore, D.~Murray, C.~Olah, M.~Schuster, J.~Shlens,
  B.~Steiner, I.~Sutskever, K.~Talwar, P.~Tucker, V.~Vanhoucke, V.~Vasudevan,
  F.~Vi\'{e}gas, O.~Vinyals, P.~Warden, M.~Wattenberg, M.~Wicke, Y.~Yu, and
  X.~Zheng, ``{TensorFlow}: Large-scale machine learning on heterogeneous
  systems,'' 2015, software available from tensorflow.org. [Online]. Available:
  \url{https://www.tensorflow.org/}
\BIBentrySTDinterwordspacing

\end{thebibliography}

\end{document}